\documentclass[letterpaper]{article} 
\usepackage{aaai24}  
\usepackage{times}  
\usepackage{helvet}  
\usepackage{courier}  
\usepackage[hyphens]{url}  
\usepackage{graphicx} 
\usepackage{natbib}  
\usepackage{caption} 
\usepackage{algorithm}
\usepackage{algorithmic}
\usepackage{svg}
\usepackage{amsmath}
\DeclareMathOperator*{\argmin}{argmin} 
\usepackage{makecell}

\usepackage{newfloat}
\usepackage{listings}
\usepackage{booktabs}
\DeclareCaptionStyle{ruled}{labelfont=normalfont,labelsep=colon,strut=off} 
\floatstyle{ruled}
\newfloat{listing}{tb}{lst}{}
\floatname{listing}{Listing}
\title{Text2Control3D: Controllable 3D Avatar Generation in Neural Radiance Fields using Geometry-Guided Text-to-Image Diffusion Model}
\author{
    Sungwon Hwang, Junha Hyung, Jaegul Choo
}
\affiliations{
    Graduate School of Artificial Intelligence, KAIST \\

    \{shwang.14, sharpeeee, jchoo\}@kaist.ac.kr
}

\usepackage{bibentry}

\begin{document}

\maketitle

\begin{abstract}
Recent advances in diffusion models such as ControlNet have enabled geometrically controllable, high-fidelity text-to-image generation. However, none of them addresses the question of adding such controllability to text-to-3D generation. In response, we propose Text2Control3D, a controllable text-to-3D avatar generation method whose facial expression is controllable given a monocular video casually captured with hand-held camera. Our main strategy is to construct the 3D avatar in Neural Radiance Fields (NeRF) optimized with a set of controlled viewpoint-aware images that we generate from ControlNet, whose condition input is the depth map extracted from the input video. When generating the viewpoint-aware images, we utilize cross-reference attention to inject well-controlled, referential facial expression and appearance via cross attention. We also conduct low-pass filtering of Gaussian latent of the diffusion model in order to ameliorate the viewpoint-agnostic texture problem we observed from our empirical analysis, where the viewpoint-aware images contain identical textures on identical pixel positions that are incomprehensible in 3D. Finally, to train NeRF with the images that are viewpoint-aware yet are not strictly consistent in geometry, our approach considers per-image geometric variation as a view of deformation from a shared 3D canonical space. Consequently, we construct the 3D avatar in a canonical space of deformable NeRF by learning a set of per-image deformation via deformation field table. We demonstrate the empirical results and discuss the effectiveness of our method.


\end{abstract}


\section{Introduction}

Recently, large text-to-image diffusion models \cite{ramesh2022hierarchical, saharia2022photorealistic, rombach2021highresolution} have enabled visually-appealing image generation that faithfully reflects the semantics of conditional texts. Such works have greatly impacted the generative models to be widely used for non-expert users, as the models can generate high-quality images of interest using text descriptions only. Moreover, ControlNet \cite{zhang2023adding} further extended the utility of the models by proposing the method to add geometric control such as canny edge, human pose, depth map, and normal map for geometrically controlled text-conditional image generation. 

Meanwhile, Text-to-3D generation methods such as DreamFusion \cite{poole2022dreamfusion} have been proposed using Neural Radiance Fields \cite{mildenhall2020nerf} as a 3D representation, and proved to yield promising quality of generation. However, no work has tackled the question of adding geometric controllability to text-to-3D generation, despite its importance highlighted by ControlNet and its impact to the research community.

Parallel to ControlNet extracting geometric control factors from a source image for generative conditions, we propose a method to generate text-conditional 3D facial avatars conditioned with geometric control factors extracted from a casually-captured monocular video of a face. As a pioneering work, we focus on controllability in terms of the facial expressions and shapes for 3D avatar generation, which are the key functionalities of an avatar to behave as a faithful, virtual agent. Specifically, given a text description and a casually-captured monocular video for conditional control, our method generates a 3D avatar that reflects the semantics of the text as well as facial expression and deformation of the source face as in Fig.~1. 

Our main strategy comprises the following procedures. First, we generate a set of viewpoint-aware images of an avatar, whose facial expressions and shapes are controlled using geometry-conditional text-to-image diffusion model such as ControlNet. Then, the generated images are used to construct the avatar in NeRF. In doing so, we define three problems raised and solved by our method, which as a whole yields the state-of-the-art 3D avatar generation quality and controllability compared to existing works. 

When leveraging ControlNet to generate viewpoint-aware images, injecting constant appearance and desired facial expression across the images is one key objective for well-controlled avatar generation. To do so, we propose cross-reference attention, a zero-shot method for the diffusion model where viewpoint-aware image generations conduct a cross-attention to a shared, referential information of controlled facial expression and appearance.

Another objective is to ameliorate the texture-sticking problem, the phenomena where images generated with slight spatial variations of latent contain constant, spatial variation-agnostic textures \cite{Karras2021}. We observed that our viewpoint-aware images conditioned with slightly varying geometric conditions also suffer from similar problem. Such disrupts high-quality 3D avatar construction, as sticking textures are not geometrically interpretable and becomes the key factor of unwanted artifacts. In this work, we show that the sticking textures are originated from high-frequency components of the Gaussian latent. From the observation, we propose a method to remove the sticking textures by filtering high-frequency components of the Gaussian latent in Fourier domain, and re-injecting texture-sticking-free details via cross-attention to full-frequency features. 

The final objective of our method is to construct 3D avatar given the viewpoint-aware images that are not strictly consistent in geometry. Our insight is to interpret such inconsistencies as views of per-image deformations from a canonical 3D model. Inspired by recent advancements in deformable NeRFs \cite{park2021nerfies, park2021hypernerf}, we train deformation field table, a set of per-image 3D deformation code, and construct a shared, 3D canonical model as our final 3D avatar.



In summary, our work makes four major contributions:

\begin{itemize}
    \item From the best of our knowledge, Text2Control3D is the first controllable Text-to-3D avatar generation method. 
    \item We propose a zero-shot method of ControlNet that generates a set of viewpoint-aware avatar images with constant appearance and controlled facial expression.
    \item We observe and ameliorate the texture-sticking problem in viewpoint-aware images generated from ControlNet.
    \item We propose a method to reconstruct high-fidelity 3D avatar given a set of generated viewpoint-aware images.
\end{itemize}

\section{Related Works}

\noindent \paragraph{Neural Radiance Fields} First proposed by \cite{mildenhall2020nerf}, Neural Radiance Field (NeRF) is an implicit representation of 3D space, which is an MLP that predicts a color and density of 3D points. These are then rendered with volume rendering techniques and supervised with dense set of posed images for optimization. Following works include dynamic scene reconstruction with deformable NeRF \cite{pumarola2021d, park2021nerfies, park2021hypernerf}, or latent-conditional NeRF for generative 3D models using GANs \cite{chan2021pi, chan2022efficient, an2023panohead}. 

\noindent \paragraph{Text-to-2D/3D Diffusion Models} Recent advancements in text-to-image diffusion models such as DALL-E \cite{ramesh2021zero}, Imagen \cite{saharia2022photorealistic}, and Latent Diffusion Models \cite{rombach2021highresolution} have brought great advancements in text-driven high-fidelity image generations. Following works such as GLIGEN \cite{li2023gligen} or ControlNet \cite{zhang2023adding} propose a method to generate images that satisfy geometric conditions such as depth, normal, or key-point maps.

Meanwhile, a line of work in 3D generation leverages distillation from large-scale text-to-image diffusion model to 3D via per-text optimization in NeRF. Notably, Score Distillation Sampling \cite{poole2022dreamfusion} and their variants \cite{Lin_2023_CVPR, chen2023fantasia3d} approximate the gradients of parameters of 3D representations with reverse diffusion guidance of large-scale text-to-image diffusion model. 
Other methods include image-to-3D generation \cite{liu2023one, liu2023zero, anciukevivcius2023renderdiffusion}, which can conduct text-to-3D generation given a text-to-image diffusion model, or diffusion models in 3D space \cite{shue20233d, luo2021diffusion}, which require 3D assets for training data.

\begin{figure*}[htbp]
  \includegraphics[width=\linewidth]{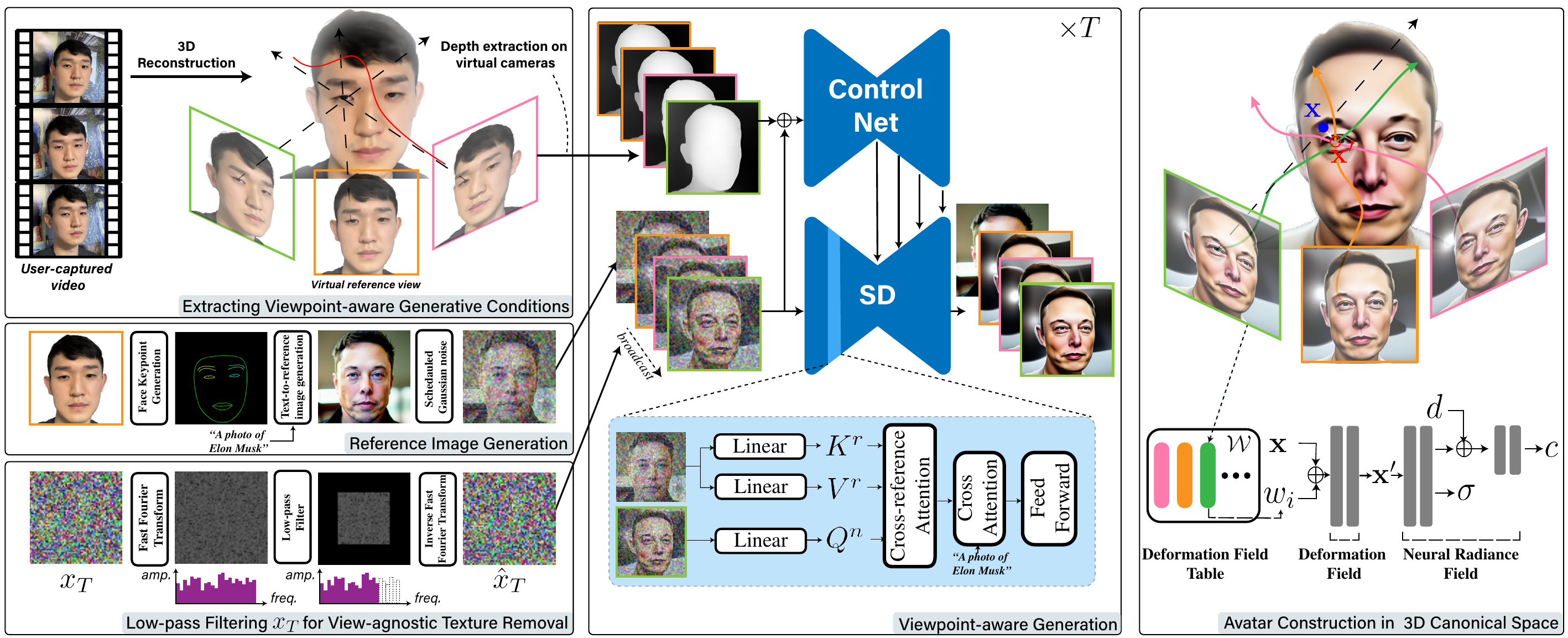}
  \caption{Illustration of Text2Control3D, our controllable text-to-3D avatar generation method.}
  \label{fig:2}
  \vspace{-0.3cm}
\end{figure*} 

\section{Proposed Method}
\label{sec:proposed}
In this section, we propose a pipeline to generate a 3D face avatar given a text description $c_{text}$, (e.g. \textit{"Elon Musk", "A handsome white man with short-fade haircut"}), and frames of a monocular video $\mathcal{I}=\{I_1 \cdots I_L\}$ for facial expression and shape control.

To do so, we first extract viewpoint-augmented depth maps from a monocular video. Then, we use these depth maps as conditions for generating viewpoint-aware avatar images using ControlNet enhanced with \textit{(1)} cross-reference attention to achieve controllability on facial expression and appearance across the viewpoint-aware generations, and \textit{(2)} low-pass filtering of Gaussian latent to remove view-agnostic textures that break 3D consistency. Finally, we regard remaining inconsistencies in geometry as views of per-image deformation of 3D canonical space. As so, we train deformation field table that encodes per-view deformation code to finally construct the 3D avatar in NeRF canonical space. The summary is illustrated in Fig.~\ref{fig:2}.

\subsection{Viewpoint-aware Generative Conditions}
First, we create a set of generative conditions from monocular video frames. We use depth maps for these conditions, as we empirically learned that they contain sufficient viewpoint knowledge while preserving reasonable amount of source geometry information that do not collide with semantics of text conditions. Instead of retrieving depth maps from input video frames, we use these frames to reconstruct the scene, followed by rendering depth maps from virtual cameras as means of augmenting to denser depth map construction.

Specifically, we first reconstruct the source face into 3D in NeRF \cite{mildenhall2020nerf} using the input monocular video frames. That is, we find

\begin{equation}
    \theta^{*}_{src} = \argmin_{\theta_{src}} \sum_{l=1}^{L} || I_l - g(\theta_{src}; T^{src}_l) ||_{2},
\end{equation}

\noindent via gradient-descent optimization, where $\theta_{src}$ is the parameter of NeRF MLP,  $g(\cdot)$ is a volume rendering function \cite{mildenhall2020nerf}, $T^{src}_l \in \mathcal{T}^{src}$ is a camera parameter of a frame $I_l$.

Using $\theta^{*}_{src}$, we then render depth maps $\mathcal{D} = \{D_1 \cdots D_N\}$ on a set of augmented virtual cameras $\mathcal{T}=\{T_1 \cdots T_N\}$ as

\begin{equation}
    D_n = g_{d}(\theta^{*}_{src}; T_n),
\end{equation}

\noindent where $D_n \in \mathcal{D}$, $T_n \in \mathcal{T}$, and $g_d$ is a volumetric depth rendering function, or a monocular depth estimation on correspondingly rendered images.

\subsection{Viewpoint-aware Image Generation}
Given our viewpoint-aware generative conditions $\mathcal{D}$ and $c_{text}$, a vanilla method to generate viewpoint-aware images of an avatar $\mathcal{X} = \{x^1 \cdots x^N\}$ using depth-conditional ControlNet $G_d$ would be

\begin{equation}
    x^n = G_{d}(D_n, c_{text}, x^{n}_{T}),
    \label{eq:controlnet_depth}
\end{equation}

\noindent where $x_{T} \sim \mathcal{N}(0, 1)$ and $x_{T} \rightarrow x^{n}_T$ for all $n$. However, we make empirical analysis of Eq.~\eqref{eq:controlnet_depth} and learned that it yields sub-optimal quality for viewpoint-aware generation. As so, we propose two methods, cross-reference attention and low-pass filtering of Gaussian latent, which are zero-shot methods to ControlNet.

\subsubsection{Cross-reference Attention} We empirically learned and reported following qualitative results in the first column of Fig.~\ref{fig:ablation}-(a) that per-frame avatar image generation using Eq.~\eqref{eq:controlnet_depth} has two undesirable properties for viewpoint-aware avatar image generation. First, depth condition alone cannot reliably convey detailed facial expressions. Second, generated images conditioned with depth images of far-distanced virtual cameras vary in their visual appearances, even though they are generated from the same Gaussian latent. 


As so, we propose cross-reference attention, a method that induces reverse diffusion processes to attend to a shared information of controlled facial expression and appearance. Specifically, we first extract a desired face key-point from the image of the source rendered at a virtual reference view. Then, we generate a reference image $\bar{x}^r$ using the face key-point, $c_{text}$ and another key-point conditional ControlNet $G_{kp}$. Formally,
    
\begin{align}
    k^r = K(g(\theta^{*}_{src}; T^r)) \\
    \bar{x}^r = G_{kp}(k^r, c_{text}), 
\end{align}

\noindent where $T^r \in \mathcal{T}$ is a virtual reference camera, and  $K(\cdot)$ is a face key-point estimator.

Then, we devise the feature maps in our viewpoint-aware generator $G_d$ to attend to the reference image via cross attention as  

\begin{gather}
    Q^n = MLP_{Q}(h^n_{t}), \\
    K^r, V^r = MLP_{K, V}(\bar{h}^{r}_{t}),
\end{gather}

\noindent where $h^n_t$ is the feature map of $x^n_t$, and $\bar{h}^{r}_{t}$ is the feature map of $\bar{x}^{r}_{t}  \sim \mathcal{N}(\sqrt{\bar{\alpha}_{t}}\bar{x}_r, (1 - \bar{\alpha}_{t})\textbf{I})$, the reference image noised with scheduled step of $t$ \cite{ho2020denoising}. Effectively, we may replace the self-attention modules to our cross-reference attention.

\begin{equation}
     \text{Attention}(Q^{n}, K^r, V^r) = softmax(\frac{Q^{n}(K^r)^{\mathsf{T}}}{\sqrt{c}})V^{r}.
     \label{eq:attention}
\end{equation}

\begin{figure}
  \includegraphics[width=\linewidth]{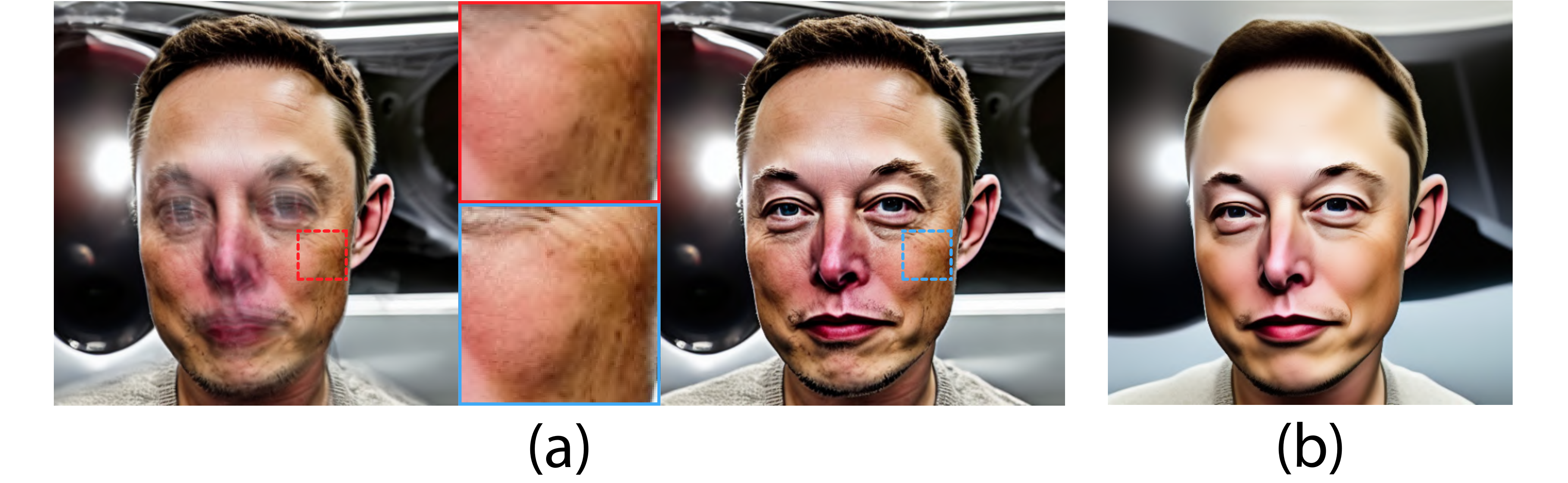}
  \caption{(a) Empirical analysis of the texture-sticking problem observed from images generated with ControlNet conditioned with depth rendered from adjacent cameras. (b) Our method ameliorates the problem by removing the sticking textures while preserving most high-fidelity details.}
  \label{fig:ablation2}
\end{figure}

\noindent \paragraph{View-agnostic Texture Removal}
We empirically found that the generated images suffer from the texture-sticking problem \cite{Karras2021}, the phenomena that the images share constant textures in image space despite of spatial variations in generative conditions, which we dub as view-agnostic texture in our work. Specifically, we generated avatar images using vanilla ControlNet conditioned with depth maps rendered from 13 adjacent virtual cameras, and averaged the images. Left side and right-side of Fig.~\ref{fig:ablation2}-(a) are the averaged image and the image generated with the depth map rendered from a virtual camera located at the center of the 13 cameras. As can be seen, some textural details of the two images are identical, which is possible only when the textures of the most generated images are located at an identical position in image space. These eventually create undesirable artifacts when used to construct 3D avatars in later sections. As so, we propose a method to remove such view-agnostic textures while preserving most high-fidelity details, as seen in Fig.~\ref{fig:ablation2}-(b).

We learned that high-frequency components in the Gaussian latent $x_{T}$ is the key factor that generates such textures. As so, we first remove the high-frequency components of the Gaussian latent in Fourier domain as

\begin{equation}
    \hat{x}^{n}_{T} = f^{-1}(P(f(x^{n}_{T}), b)),
    \label{eq:low_pass}
\end{equation}

\noindent where $f$ and $f^{-1}$ are Fast Fourier Transform and Inverse Fast Fourier transform \cite{brigham1988fast}, respectively, $P$ is a low-pass filter that removes the frequency components higher than $b$, a cut-off frequency.

Then, high-fidelity details are re-injected via cross-attention to full-frequency features as  

\begin{gather}
    Q^{n} = MLP_{q}(\hat{h}^{n}_{t}), \\
    K^n, V^n = MLP_{K, V}(h^{n}_{t}),
\end{gather}

\noindent where $\hat{h}^{n}_{t}$ is the feature map of $\hat{x}^{n}_{t}$.
In this case, $h^{n}_{t}$ is the feature map of $x^{n}_{t}$ that goes through an independent reverse diffusion process from $x_T$ in parallel. Design choice of cross-attention to full-frequency features instead of self-attention will be discussed in ablation study. 

Finally, we merge both cross-reference attention and view-agnostic texture removal into a single attention pipeline by replacing $h^{n}_{t}$ in Eq.~(11) with $\bar{h}^{r}_{t}$ as

\begin{gather}
    Q^{n} = MLP_{q}(\hat{h}^{n}_{t}),\label{eq:query_final} \\
    K^r, V^r = MLP_{K, V}(\bar{h}^{r}_{t}).\label{eq:key_value_final}
\end{gather}

Formally, Eq.~\eqref{eq:controlnet_depth} is reformulated to define our final viewpoint-aware image generator as

\begin{equation}    
    x^n = G^{*}_{d}(D_n, D_r, c_{text}, \hat{x}^{n}_T, \bar{x}^r),
    \label{eq:final_controlnet_depth}
\end{equation}
\noindent where $G^{*}_{d}$ replaces self-attention with our cross-reference attention that comprises Eq.~\eqref{eq:query_final}, Eq.~\eqref{eq:key_value_final}, and Eq.~\eqref{eq:attention}, and $D_r$ is depth map rendered from a virtual reference camera using $\theta^{*}_{src}$, which is used as a depth condition for calculating $\bar{h}^{r}_{t}$. In addition, cross-reference attention is conducted on time-steps larger than $t_{thres}$ to regularize facial expression and appearances on earlier stages, whereas self-attention replaces the cross-reference attention on later stages to concentrate on viewpoint-specific generations.



\begin{figure*}[!ht]
  \includegraphics[width=\linewidth]{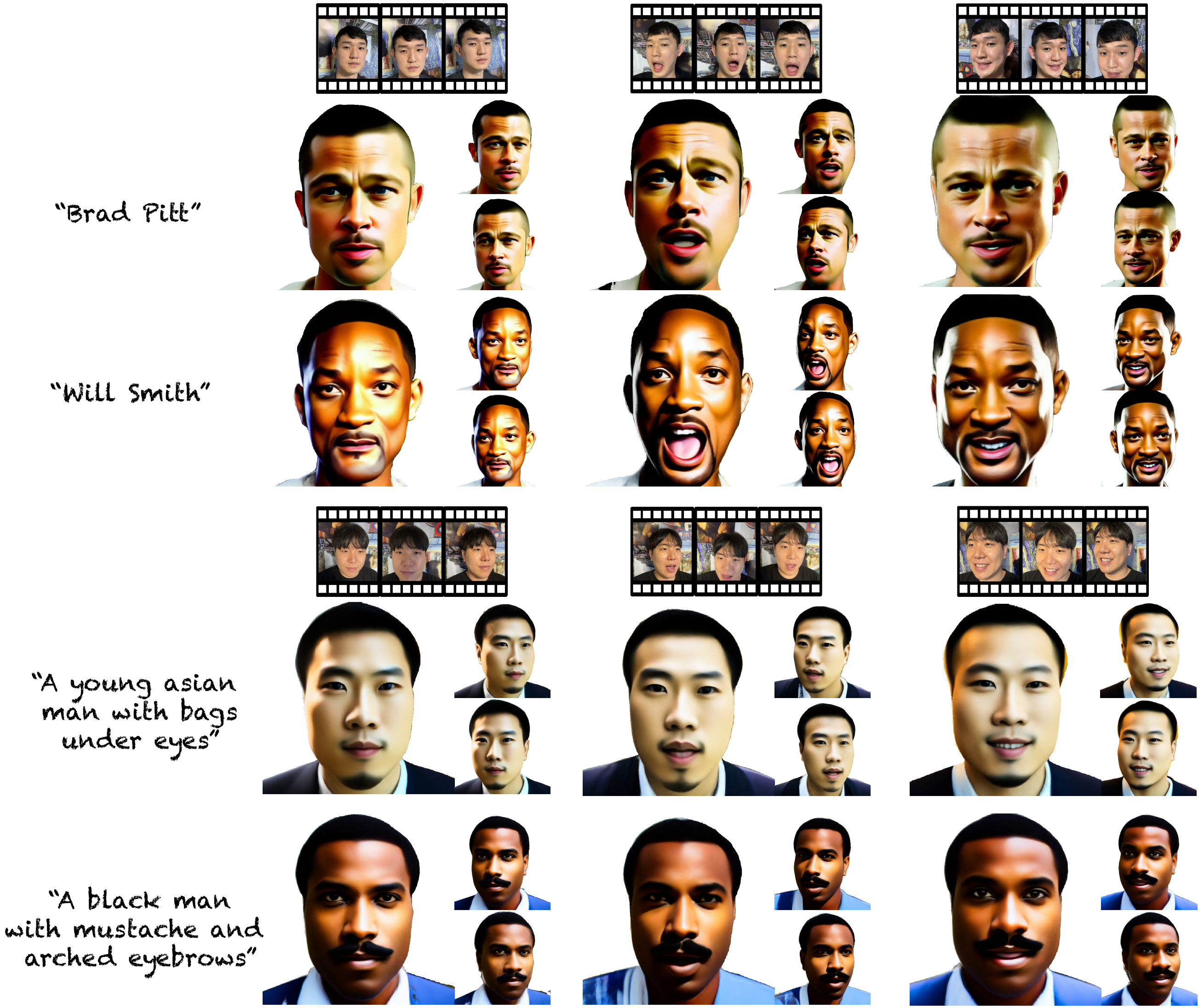}
  \caption{Visualization of 3D avatars generated with our method.}
  \label{fig:result}
  \vspace{-0.5cm}
\end{figure*}

\subsection{3D Avatar Construction in Canonical Space}
Using $\mathcal{X}$, a set of generated viewpoint-aware avatar images, we now reconstruct a 3D avatar model in NeRF. 
Since the generated images are not strictly consistent in geometry, direct reconstruction may yield broken results and unwanted artifacts. Instead, we assume that all images share a single 3D canonical model, and regard per-image geometric inconsistency as a view of a deformed canonical model. 

Specifically, we first define learnable deformation field table, $\mathcal{W} = \{w_1 \cdots w_N \}$, where $w_n \in \mathcal{W}$ is assigned to each virtual view. Then, we define a canonical 3D model parameter $\Theta = \{\theta_{d}, \theta_{c}\}$, where each deformation code $w_n$ learns to project a point in observation space $\textbf{x}$ to the shared canonical space, after which a shared MLP predicts the view-direction dependent color and density of the point as
\begin{gather}
    \textbf{x}' = F(\textbf{x}; \theta_{d}, w_{n}), \\
    (c, \sigma) = F'(\textbf{x}', d; \theta_{c}),
\end{gather}

\noindent where $F$ and $F'$ are MLP forward functions, and $d$ is a view direction in observation space. We optimize $\Theta$ and $\mathcal{W}$ to minimize the difference between the generated avatar images $\mathcal{X}$ and images rendered from our deformed canonical space. Also, we do not reconstruct backgrounds in generated images, as they are not viewpoint-aware and are irrelevant to avatar reconstruction. As so, we use off-the-shelf face segmentation network to generate a set of per-image mask $M_n \in \mathcal{M} = \{M_1 \cdots M_N\}$, and apply sparsity loss, which encourages the point samples of camera rays projected to background region to yield low density, thus emptying the corresponding space. 

Also, we learned that the canonical NeRF occasionally converges to semi-transparent object density in order to explain the geometric inconsistencies between viewpoint-aware images
As so, we apply entropy loss \cite{Kim_2022_CVPR}, which maximizes the Shannon entropy \cite{shannon1948mathematical} of the density distribution of a ray projected to foreground. Such prior encourages the object to maximize the density closer to the object surface and to suppress the density farther from the object surface, thus discouraging semi-transparent surfaces of 3D canonical avatar. In summary, we minimize the following loss for 3D avatar reconstruction, $\mathcal{L}_{NeRF} = \lambda_{RGB}\mathcal{L}_{RGB} + \lambda_{sp}\mathcal{L}_{sp} + \lambda_{entropy}\mathcal{L}_{entropy}$, where

\begin{gather}
    \mathcal{L}_{RGB} = \sum_{n} || M_n \otimes (x^n - g(\Theta, w_n; T_n)) ||_{2}, \\
    \mathcal{L}_{sp} = \frac{1}{J}\sum_{n, j, h, w}|1-\text{exp}(-\lambda \sigma_{j})| \cdot (1 - m^{(h, w)}_n), \\
    \mathcal{L}_{entropy} = -\sum_{n, j, h, w} p(\alpha_j)\text{log} p(\alpha_j)  \cdot m^{(h, w)}_n,
\end{gather}

\noindent where $m^{(h, w)}_n$ is a binary value of $M_{n}$ located at $(h, w)$ pixel location, $\{\sigma_{j}\}_{j=1}^{J}$ are density values evaluated from points sampled from the ray propagating from $(h, w)$ of image plane located at $T_n$, $p(\alpha_j) = \frac{\alpha_{j}}{\sum_{i=1}^{J}\alpha_{i}} = \frac{1-\text{exp}(-\sigma_{j}\delta_{j})}{\sum_{i=1}^{J}1-\text{exp}(-\sigma_{i}\delta_{i})}$ is the normalized point opacity, and $\lambda_{RGB}, \lambda_{sp}, \lambda_{entropy}$ are hyper-parameters.

\begin{figure}
  \includegraphics[width=\linewidth]{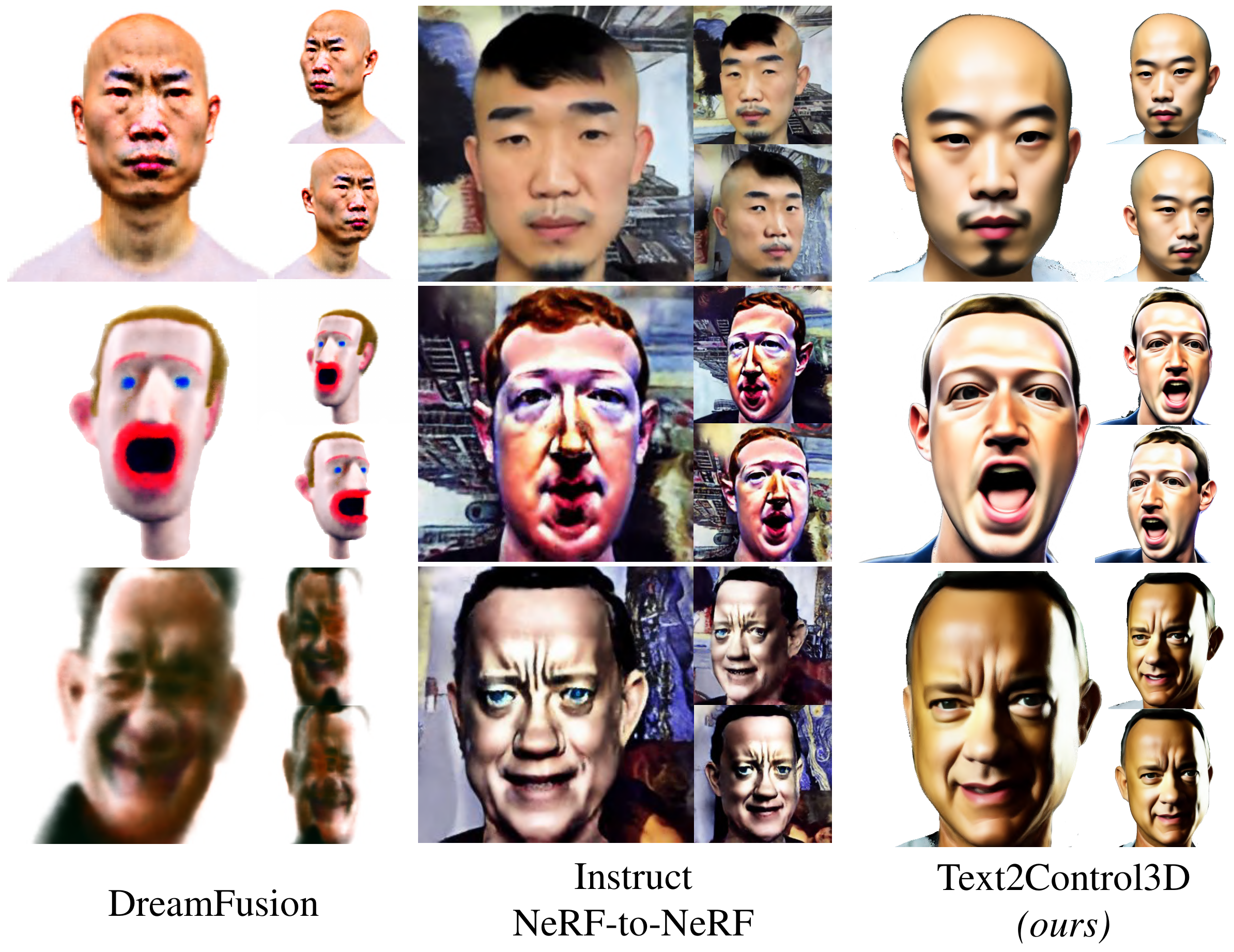}
  \caption{Qualitative comparisons to baselines. Text conditions are \textit{"A bald Asian man with goatee"}, \textit{"Mark Zuckerberg"}, and \textit{"Tom Hanks"} for each row from the top.
  }
  \label{fig:comparison}
  \vspace{-0.3cm}
\end{figure}

\section{Experiments}

\paragraph{Dataset} Volunteers were instructed to make three distinct and extreme facial expressions: neutral face, smiling, and opened mouth, in order to validate the range of expressibility. For text conditions, we created two sets of text prompts, each containing 23 number of texts: one is a set of names of celebrities, and the other is a set of descriptions of facial attributes. 

\paragraph{Implementation Details} We generate view-point aware images in $512 \times 512$ resolution. ControlNet conducts reverse diffusion process in latent space defined by pre-trained auto-encoder \cite{rombach2021highresolution}, making spatial dimension of Fourier spectrum to be $64 \times 64$, which is the latent space resolution. We use $b=22$ for attribute description texts, and $b=18$ for celebrity name texts for low-pass filtering in Fourier space. Algorithmically, low-pass filter $P$ replaces the values in Fourier spectrum outside the $(2\cdot b) \times (2\cdot b)$ box located at the spectrum center with 0. For reverse diffusion process of ControlNet, we used DDIM \cite{song2020denoising} with 50 steps, and $t_{thres}=45$ for cross-reference attention. For canonical 3D avatar reconstruction, we used $\lambda_{RGB}=1$, $\lambda_{sp}=1 \times 10^{-3}$, and $\lambda_{sp}=1 \times 10^{-6}$. 

\paragraph{Baselines} Since no work tackles the controllable text-to-3D generation from the best of our knowledge, we make qualitative comparisons to DreamFusion \cite{poole2022dreamfusion}, a text-to-3D generation model, and Instruct-NeRF2NeRF \cite{instructnerf2023}, a NeRF to NeRF translation method using InstructPix2Pix. For DreamFusion, we concatenate additional texts that describe a desired facial expression. 

\section{Results and Discussion}

\noindent \paragraph{Qualitative Results} We report the qualitative results of our method in Fig.~1 and Fig.~\ref{fig:result}. The generated 3D avatars faithfully reflects the semantics of the text descriptions, facial expression and the coarse shape of the face in the source monocular video. In addition, despite that contour of the source face remains in the avatars, their unique geometries imply that ControlNet can flexibly handle the depth map conditions to generate images with novel shapes while preserving the viewpoint-aware information implicitly encoded in depth conditions.

\begin{table}
  \centering
  \resizebox{\linewidth}{!}{\begin{tabular}{@{}cccc@{}}
    \toprule
    & Visual Fidelity $\uparrow$ & Text Reflectivity $\uparrow$ & Structural Reflectivity $\uparrow$ \\
    \midrule
    DreamFusion & 2.11 & 2.68 & 2.88 \\
    Instruct-NeRF2NeRF & 3.31 & 3.82 & 3.23 \\
    Text2Control3D & \textbf{4.92} (\textit{+1.61}) & \textbf{4.55} (+\textit{0.73}) & \textbf{4.87} (+\textit{1.64}) \\

    \bottomrule
  \end{tabular}}
  \caption{User study results. \textbf{Best results} are highlighted in bold, and \textit{difference to the second best} is italicized.}
  \label{tab:user_study}

  \centering
   \resizebox{\linewidth}{!}{\begin{tabular}{@{}cccc@{}}
    \toprule
    \textcolor{white}{aaaa} & DreamFusion & Instruct-NeRF2NeRF & Text2Control3D \\
    \midrule
    \textcolor{white}{aaa} R-Precision $\uparrow$ \textcolor{white}{aaa} & 65.2 & 56.5 & \textbf{69.5} \\
    \bottomrule
  \end{tabular}}
  \caption{Evaluating R-Precision using CLIP ViT-L/14. All images are rendered from novel camera views.}
  \vspace{-0.5cm}
  \label{tab:rprec}

\end{table}

\noindent \paragraph{Comparison to Baselines} The following qualitative comparisons are reported in Fig.~\ref{fig:comparison}. As can be seen, the 3D avatars generated from our method shows the highest fidelity and controllability of facial expressions. Since there is no ground truth images to measure PSNR \cite{mildenhall2020nerf} or 3D shape to measure Chamfer Distance \cite{park2019deepsdf}, we heavily rely on human evaluations for comparisons. Specifically, we asked evaluators to score the quality of results in three criteria: visual fidelity in terms of the overall shape, texture, and color of the faces, text reflectivity in terms of how well the generated avatar is faithful to semantics of the text conditions, and structural reflectivity in terms of how well the generated avatar is faithful to facial expression conditions. We report the average of the responses in Table.~\ref{tab:user_study}. As can be seen, results from our methods were preferred by those from baselines in all criteria. 

Following prior text-to-3D generation works \cite{poole2022dreamfusion}, we also computed R-Precision in CLIP ViT-L/14 latent space. Specifically, R-precision measures the percentage of correct retrieval of text that generated the 3D avatar among a set of incorrect texts given the image rendering of the 3D avatar. Our method showed reasonable performance compared to the baselines as reported in Table.~\ref{tab:rprec}. 


DreamFusion yielded the lowest scores on all user study criteria, inferring lower visual quality than other methods. Occasionally, the 3D shape is also constructed unnaturally as in the third row result. Along with the prevalent Janus problem \cite{armandpour2023re} observed from SDS-based methods, where frontal appearances are created on side and rear parts of 3D object, these are evidences that SDS with text-to-image diffusion model is not strictly viewpoint-aware, making it unsuitable for sophisticated geometry control for 3D generation. 

Instruct-NeRF2NeRF yields reasonable translation results, yet user study responses indicate inferior results than our method. In addition, translation was not successful in case of the first row result, where hair is not completely removed to translate the source head into bald.  

\begin{table}
  \resizebox{\linewidth}{!}{\begin{tabular}{@{}ccccc@{}}
    \toprule
    & Method & RMSE $\downarrow$ &  PCK $\uparrow$ &  CFS $\uparrow$ \\
    \midrule
    \textit{(a)} & \makecell{$Q = MLP_{Q}(h^{n}_{t})$ \\ $K, V = MLP_{K, V}(h^{n}_{t})$} & 53.2 & 78.4 & 0.682 \\
    \midrule
    \textit{(b)} & \makecell{$Q = MLP_{Q}(\hat{h}^{n}_{t})$ \\ $K, V = MLP_{K, V}(\hat{h}^{n}_t)$} & 72.3  & 60.7 & 0.710 \\
    \midrule
    \textit{(c)} & \makecell{$Q = MLP_{Q}(\hat{h}^{n}_{t})$ \\ $K, V = MLP_{K, V}(h^{n}_t)$} & 53.2 & 78.4 & 0.684 \\
    \midrule
    \textit{(d)} & \makecell{$Q = MLP_{Q}(\hat{h}^{n}_{t})$ \\ $K, V = MLP_{K, V}(\bar{h}^{r}_{t})$} & \textbf{39.5}  & \textbf{84.2}  & \textbf{0.738} \\
    \bottomrule
  \end{tabular}}
  \caption{Ablation study on the controllability of facial expression and consistency of appearance over viewpoints.}
  \vspace{-0.5cm}
  \label{tab:ablation}
\end{table}

\begin{figure}
  \includegraphics[width=\linewidth]{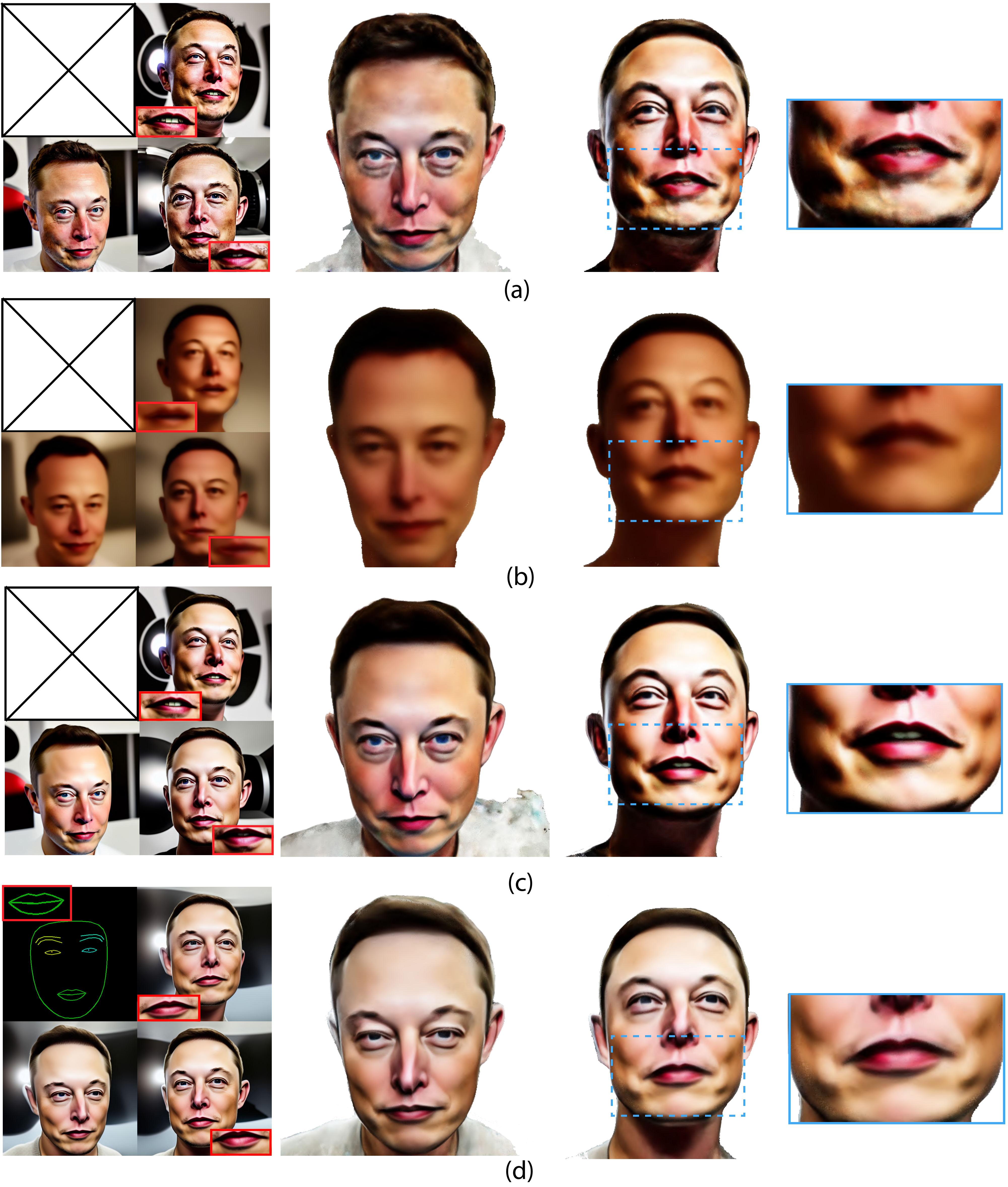}
  \caption{Ablation results. Each row corresponds to modifications in Table.~\ref{tab:ablation}. First column shows the viewpoint-aware images and facial key-points if used.}
  \vspace{-0.5cm}
  \label{fig:ablation}
\end{figure}

\noindent \paragraph{Ablations} 
Table~\ref{tab:ablation} and Fig.~\ref{fig:ablation} show ablation on cross-frame attention and low-pass filtering of Gaussian latents to study their impacts on viewpoint-agnostic texture problem, controllability on facial expressions and appearance across viewpoint-aware images. For facial expression controllability we measure Root Mean Squared Error (RMSE) and Percentage of Correct Key-points (PCK) \cite{li2020cascaded} between the face key-points of the source and those from the generated avatars, all of which are extracted from images rendered from novel camera views. Consistency in appearance are measured with Cosine Face Similarity (CFS) between avatar images rendered from novel views and an image rendered from a reference view using the off-the-shelf face recognition model\footnote{https://github.com/ronghuaiyang/arcface-pytorch}.

We ablate four major modifications of U-Net in ControlNet: \textit{(a)} vanilla self-attention, \textit{(b)} low-pass filtering of Gaussian latent with self-attention, \textit{(c)} cross-attention of low-pass filtered Gaussian latent features to full-frequency feature that goes through reverse diffusion process in parallel, i.e. Eq.~(10) and Eq.~(11), and \textit{(d)} cross-attention of features de-noised from low-pass filtered Gaussian latent to reference image with scheduled noise, which is the method we suggest. First column in Fig.~\ref{fig:ablation} are correspondingly generated viewpoint-aware images, and the rest of the columns are images of reconstructed avatars rendered from novel views.  

Without low-pass filtering \textit{(a)} creates viewpoint-agnostic textures that create unwanted artifacts in 3D avatar. However, \textit{(b)} yields over-smoothed result without cross-attention of low-pass filtered features to full-frequency features. Without cross-attention to a reference image as in \textit{(a)}, \textit{(b)} and \textit{(c)}, viewpoint-aware images are not coherent in appearance and facial identity, which forces the NeRF model to over-fit to varying appearances via view-dependence of the MLP network. Also, inconsistent facial expressions in viewpoint-aware images in \textit{(a)}, \textit{(b)}, and \textit{(c)} induce NeRF to reconstruct an under-controlled shape. For example in Fig.~\ref{fig:ablation}-(a-c), mouths are observed to be closed from the upper view, but are in fact slightly opened when observed from lower views. 

\section{Limitation}
One limitation of our method is that key-point conditional ControlNet constructs the controllability bottleneck of our method. For instance, ControlNet often fails to generate reference images with facial expressions that are relatively less common in in-the-wild training images such as closed, winked, frowning eyes. Thus, improvements in controllability of ControlNet via future research can bring parallel improvements to our pipeline in terms of the scope of facial expressions for controlled avatar generation.

\section{Conclusion}
We have presented Text2Control3D, the first work to address a controllable text-to-3D avatar generation using text-to-image diffusion model from the best of our knowledge. Our work makes three major contributions: cross-frame attention for viewpoint-aware image generation with controlled facial expression and appearance, low-pass filtering Gaussian latent that ameliorates texture-sticking problem in conditional text-to-image generation, and employing deformable NeRF to reconstruct a 3D avatar using viewpoint-aware yet geometrically inconsistent images. Text2Control3D outperformed the baselines in user-study and quantitative comparisons.

\bibliography{aaai24}

\end{document}